# Extension of Max-Min Ant System with Exponential Pheromone Deposition Rule

Ayan Acharya [#1], Deepyaman Maiti[#2], Aritra Banerjee[#3], R. Janarthanan[*4], Amit Konar [#5]

[#] *Department of Electronics and Telecommunication Engineering, Jadavpur University*
*Kolkata: 700032, India*

[1] masterayan@gmail.com
[2] deepyamanmaiti@gmail.com
[3] aritraetce@gmail.com
[5] konaramit@yahoo.co.in

[*] *Jaya Engineering College*
*C.T.H.Road, Thiruninravur, Chennai 602024, Thiruvallur District, Tamilnadu, India*
[4] srmjana_73@yahoo.com

*Abstract*— **The paper presents an exponential pheromone deposition approach to improve the performance of classical Ant System algorithm which employs uniform deposition rule. A simplified analysis using differential equations is carried out to study the stability of basic ant system dynamics with both exponential and constant deposition rules. A roadmap of connected cities, where the shortest path between two specified cities are to be found out, is taken as a platform to compare Max-Min Ant System model (an improved and popular model of Ant System algorithm ) with exponential and constant deposition rules. Extensive simulations are performed to find the best parameter settings for non-uniform deposition approach and experiments with these parameter settings revealed that the above approach outstripped the traditional one by a large extent in terms of both solution quality and convergence time.**

## I. Introduction

***Stigmergy*** is a special kind of communication prevalent among many species of ants. While roaming from food sources to the nest and vice versa, ants deposit on the ground a substance called pheromone, forming in this way a pheromone trail. Ants can smell pheromone and choose, in probability, paths marked by stronger pheromone concentration. Thus the pheromone trail allows the ants to find their way back to the food source or to the nest. Denebourg et al. [4] first studied the pheromone laying and following behavior of ants. Ant System (AS) and Ant Colony Optimization (ACO) actually owe their inspiration to the works of Denebourg et al.

The paper attempts to extend the ant system model by introducing an exponential pheromone deposition approach. We solve the deterministic ant system dynamics using differential equation. The analysis helps in determining the range of parameters in both forms of pheromone deposition rule to confirm stability in pheromone trails. The deterministic solution undertaken does not violate the stochastic nature of the Ant System algorithm since a segment of trajectory here is also selected probabilistically.

The apparent correlation between the selection of exponential pheromone deposition approach and the expected improved convergence time as well as solution quality of extended AS can be explained in the following way. A uniform pheromone deposition by an ant cannot ensure subsequent ants to follow the same trajectory. However, an exponentially increasing time function ensures that subsequent ants close enough to a previously selected trial solution will follow the trajectory, as it can examine gradually thicker deposition of pheromones over the trajectory. Naturally, ***deception probability*** ([3]) being less, convergence time should improve.

Our previous work [8] was based on stability analysis using difference equation. In this paper, we have employed differential equations which not only characterize the system dynamics more precisely but also are more popular than difference equation. The previous paper lacked sufficient experimentations to establish the betterment of the proposed deposition rule. The experiments performed over TSP instances could not at all highlight the philosophy of the non uniform deposition rule. This paper presents sufficient simulation results to establish the proposed algorithm's superiority over the traditional one. Problem environment is also chosen very cleverly to emphasize the efficacy of the proposed algorithm. Exhaustive experimentations also help find out the suitable values of parameter for which the proposed algorithm works best and from these results we attempt to ascertain an algebraic relationship between the parameter set of the algorithm and feature set of the problem environment.

The paper is divided into seven sections. Section II gives a brief description of **AS** (Ant System) and **MMAS (**MAX-MIN Ant System**).** In section III, a scheme for the general solution of Ant System is formulated. Stability analysis with closed form solution of different pheromone deposition rules is undertaken in section IV. Parameter settings for MMAS are provided in a separate module in section V. Performance analyses of the extended and classical AS are compared in section VI by using Max-Min variation of basic Ant System algorithm. Finally, the conclusions are listed in section VII.

## II. Ant System and Max-Min Ant System

Ant algorithms have largely been used in solving different **NP** hard optimization problems since their discovery. One of such applications is the Travelling Salesperson Problem (TSP) ([5]). The theory of ant system can best be explained in the context of TSP. Formally, the TSP is the problem of finding the shortest Hamiltonian circuit of a set of nodes. The basic ACO algorithm for TSP can be described as follows:

**procedure** ACO algorithm for TSPs
- Set parameters, initialize pheromone and ants' memory

**while** (termination condition not met)
- Construct Solution
- Apply Local Search ( optional)
- Best Tour check
- Update Trails

**end**

**end** ACO algorithm for TSPs

Ant System ([1],[2],[7]) was the earliest implementation of the ACO algorithm. Basically it consists of two levels:

1. **Initialization: 1.**Any initial parameters are loaded. 2. Edges are set with an initial pheromone value. 3. Each ant is individually placed on a random city.
2. **Main Loop:**

- **Construct Solution**

Each ant constructs a tour by successively applying the probabilistic choice function which can be described as follows:

$$P_i^k(j) = \begin{cases} (\tau_{ij}^{\alpha}).(\eta_{ij}^{\beta}) / \sum_{k:\, k \in N_i^k} (\tau_{ik}^{\alpha}).(\eta_{ik}^{\beta}) \text{ if } q<q_0 \\ 1 \text{ if } (\tau_{ij}^{\alpha}).(\eta_{ij}^{\beta}) = \max\{(\tau_{ik}^{\alpha}).(\eta_{ik}^{\beta}) : k \in N_i^k\} \text{ with } q>q_0 \\ 0 \text{ if } (\tau_{ij}^{\alpha}).(\eta_{ij}^{\beta}) \neq \max\{(\tau_{ik}^{\alpha}).(\eta_{ik}^{\beta}) : k \in N_i^k\} \text{ with } q>q_0 \end{cases} \quad (1)$$

where $P_i^k(j)$ is the probability of selecting node $j$ after node $i$ for ant $k$. A node $j \in N_i^k$ ($N_i^k$ being the neighborhood of ant $k$ when it is at node $i$) if $j$ is not already visited. $\eta_{ik}$ is the visibility information generally taken as the inverse of the length of link *(i,k)*, $\tau_{ik}$ is the pheromone concentration associated with the link *(i,k)*. $q_0$ is a pseudo random factor deliberately introduced for path exploration and α, β are the weights for pheromone concentration and visibility.

- **Apply Local Search:** Not used in Ant System, but is used in several variations of the TSP problem where 2-opt or 3-opt local optimizers are used.
- **Best Tour check:** Calculate the lengths of the ants' tours and compare with best tour length so far. If there is an improvement, update it.
- **Update Trails: 1.** Evaporate a fixed proportion of the pheromone on each edge. **2.** For each ant perform the ***'Ant Cycle'*** pheromone update. In ***'Ant Cycle'*** heuristic ants first complete the tour and then deposit pheromone on the entire path with an amount proportional to the inverse of the total length of the path.

Max-Min ant system ([6]) introduces four main modifications in AS. They can be highlighted as follows: **1.** To exploit the best tour found, this heuristic allows the deposition only by the iteration best ant or by the best-so-far ant. **2.** To nullify the effect of early convergence to a good but suboptimal solution due to the over exploitation of best solutions, the algorithm limits the pheromone trail values in the range $[\tau_{min}, \tau_{max}]$. **3.** The pheromone trails are initialized to upper pheromone trail limits which together with a small pheromone evaporation rate facilitates the exploration of tours at the start of the search. **4.** Finally, the pheromone trails are reinitialized each time the system moves into deadlock situation or no improved tour is generated for certain number of iterations. Detailed description of parameter settings has been provided in section 5.

## III. Deterministic Framework for Solution of Basic Ant System Dynamics

This section focuses on the development of deterministic framework using differential equation. Now, let us consider a small segment of the tour by an ant. Let $i$ and $j$ be two successive nodes, on the tour of an ant and $\tau_{ij}(t)$ be the pheromone concentration at time t associated with the edge of the graph joining the nodes $i$ and $j$.

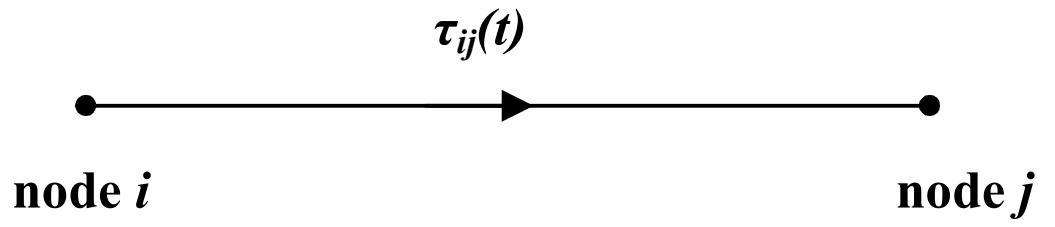

**Fig 1: Defining $\tau_{ij}(t)$**

Let $\rho>0$ be the pheromone evaporation rate and $\Delta\tau_{ij}^k(t)$ be the pheromone deposited by ant k at time t. The basic pheromone updating rule in **AS** with $m$ number of ants is therefore given by,

$$\tau_{ij}(t)=(1-\rho)\tau_{ij}(t-1)+\sum_{k=1}^{m}\Delta\tau_{ij}^k(t) \quad (2)$$

In **MMAS** algorithm, as stated in earlier section, only the best-so-far ant or the iteration-best ant is allowed to deposit pheromone on the arcs it has visited. The pheromone updating rule (2) is therefore modified as,

$$\tau_{ij}(t)=(1-\rho)\tau_{ij}(t-1)+\Delta\tau_{ij}^{bs} \quad (3)$$

$\Delta\tau_{ij}^{bs}$ in (3) is defined as $\Delta\tau_{ij}^{bs} = \begin{cases} 1/C^{bs}, & \text{if arc (i,j) belongs to } T^{bs} \\ 0, & \text{otherwise} \end{cases}$, where $T^{bs}$ is the tour conducted by either the iteration-best ant or the best-so-far ant and $C^{bs}$ is the tour length of $T^{bs}$. Therefore, in any iteration, only the arcs belonging to the best-so-far ant or the iteration-best ant receive pheromone. Now, from the pheromone update equation of Ant System i.e. from (2), it follows,

$$\tau_{ij}(t)-\tau_{ij}(t-1)=-\rho\tau_{ij}(t-1)+\sum_{k=1}^{m}\Delta\tau_{ij}^k(t)$$

$$\Rightarrow \frac{d\tau_{ij}}{dt} = -\rho\tau_{ij} + \sum_{k=1}^{m} \Delta\tau_{ij}^{k}(t)$$

$$\Rightarrow (D+\rho)\tau_{ij}(t-1) = \sum_{k=1}^{m} \Delta\tau_{ij}^{k}(t)$$

$$\therefore (D+\rho)\tau_{ij}(t) = \sum_{k=1}^{m} \Delta\tau_{ij}^{k}(t+1) \qquad (4)$$

Evidently, (4) gives the solution for the ant dynamics. Now, to solve (4), we have to separate the complimentary function and the particular integral. We now consider two different forms of $\Delta\boldsymbol{\tau}_{ij}^{k}(t)$ and try to determine the complete solution of $\boldsymbol{\tau}_{ij}(t)$.

**Evaluation of Complimentary Function (CF):**

The complimentary function of (4) is obtained by setting $\sum_{k=1}^{m} \Delta\tau_{ij}^{k}(t)$ to zero. This gives only the transient behavior of the ant system dynamics. Therefore, from (4), $(D+\rho)\tau_{ij} = 0$, $\Rightarrow D = -\rho$

Thus, the transient behavior of the Ant System is given by

CF: $\boldsymbol{\tau}_{ij}(t) = Ae^{-\rho t}$ (5)

where $A$ is a constant which is to be found out from initial condition.

**Evaluation of Particular Integral for Both Forms of Deposition Rule:**

The steady state solution of the ant system dynamics is obtained by computing particular integral of (4). This is given by,

$$\tau_{ij}(t) = \frac{1}{D+\rho} \sum_{k=1}^{m} \Delta\tau_{ij}^{k}(t+1) \qquad (6)$$

**Case I:** When $\Delta\tau_{ij}^{k}(t)=C_k$, we obtain from (6)

$$PI = \frac{1}{D+\rho}\sum_{k=1}^{m} C_k = \frac{1}{\rho}.(1+D/\rho)^{-1}\sum_{k=1}^{m} C_k$$

$$= \frac{1}{\rho}(1-\frac{D}{\rho}+\frac{D^2}{\rho^2}-.......)\sum_{k=1}^{m} C_k = \frac{1}{\rho}(1)\sum_{k=1}^{m} C_k = \sum_{k=1}^{m} C_k/\rho \qquad (7)$$

**Case II:** When $\Delta\tau_{ij}^{k}(t)=C_k(1-e^{-t/T})$, we obtain from (6),

$$\text{PI} = \frac{1}{D+\rho}\sum_{k=1}^{m} C_k(1-e^{-[t+1]/T})$$

$$= \frac{1}{D+\rho}\sum_{k=1}^{m} C_k - \frac{1}{D+\rho}\sum_{k=1}^{m} C_k e^{-(t+1)/T}$$

$$= \frac{\sum_{k=1}^{m} C_k}{\rho} - \frac{\sum_{k=1}^{m} C_k e^{-(t+1)/T}}{(\rho - \frac{1}{T})} \qquad (8)$$

## IV. Stability Analysis of Ant System dynamics with Complete Solution

In this section, we obtain the complete solution of the ant system dynamics for determining the condition for stability of the dynamics.

**Case I:** For constant deposition rule, the complete solution can be obtained by adding CF and PI from (5) and (7) respectively and is given by,

$\tau_{ij}(t) = Ae^{-\rho t} + \sum_{k=1}^{m} C_k/\rho$. At t=0,

$\tau_{ij}(0) = A + \sum_{k=1}^{m} C_k/\rho$, $\Rightarrow A = \tau_{ij}(0) - \sum_{k=1}^{m} C_k/\rho$

Therefore, the complete solution is,

$$\tau_{ij}(t) = [\tau_{ij}(0) - \sum_{k=1}^{m} C_k/\rho]\, e^{-\rho t} + \sum_{k=1}^{m} C_k/\rho \qquad (9)$$

It follows from (9) that the system is stable for ρ>0 and converges to steady state value $\sum_{k=1}^{m} C_k/\rho$ as time increases. The plot below supports the above observation.

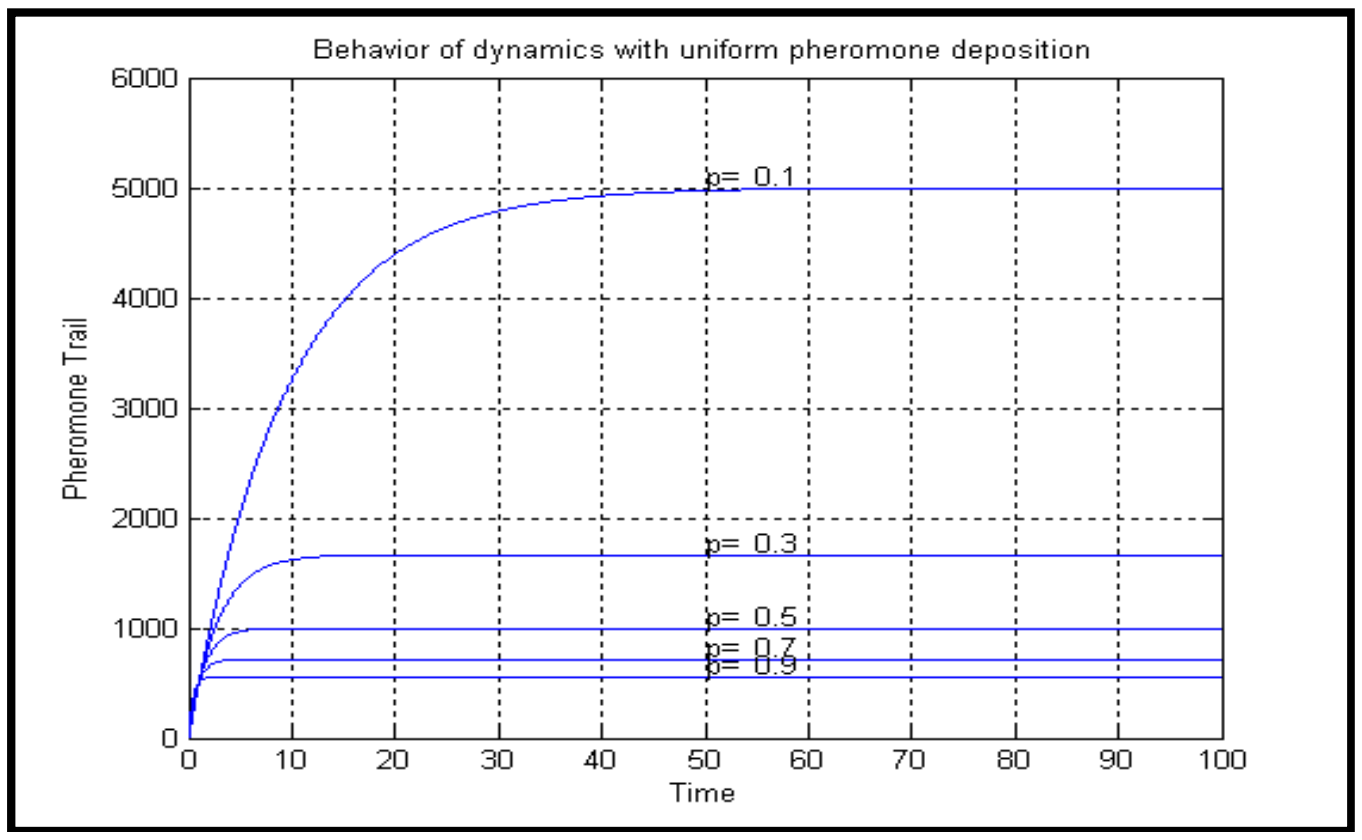

**Figure 2: $\tau_{ij}(t)$ versus t for constant pheromone deposition**

**Case II:** For exponentially increasing pheromone deposition, the complete solution is,

$$\tau_{ij}(t) = Ae^{-\rho t} + \sum_{k=1}^{m} C_k/\rho - \sum_{k=1}^{m} C_k e^{-(t+1)/T}/(\rho - \frac{1}{T})$$

Now, at t=0, $\tau_{ij}(0) = A + \sum_{k=1}^{m} C_k/\rho - \sum_{k=1}^{m} C_k e^{-(1/T)}/(\rho - \frac{1}{T})$

$$\therefore A = \tau_{ij}(0) - \sum_{k=1}^{m} C_k/\rho + \sum_{k=1}^{m} C_k e^{-(1/T)}/(\rho - \frac{1}{T})$$

$$\therefore \tau_{ij}(t) = [\tau_{ij}(0) - \sum_{k=1}^{m}\frac{C_k}{\rho} + \sum_{k=1}^{m}\frac{C_k e^{-(1/T)}}{(\rho-\frac{1}{T})}]e^{-\rho t} + \sum_{k=1}^{m}\frac{C_k}{\rho} - \sum_{k=1}^{m}\frac{C_k e^{-(t+1)/T}}{(\rho-\frac{1}{T})} \qquad (10)$$

Clearly, the system is stable for positive values of $\rho$ and T and converges to $\sum_{k=1}^{m} C_k/\rho$ in its steady state.

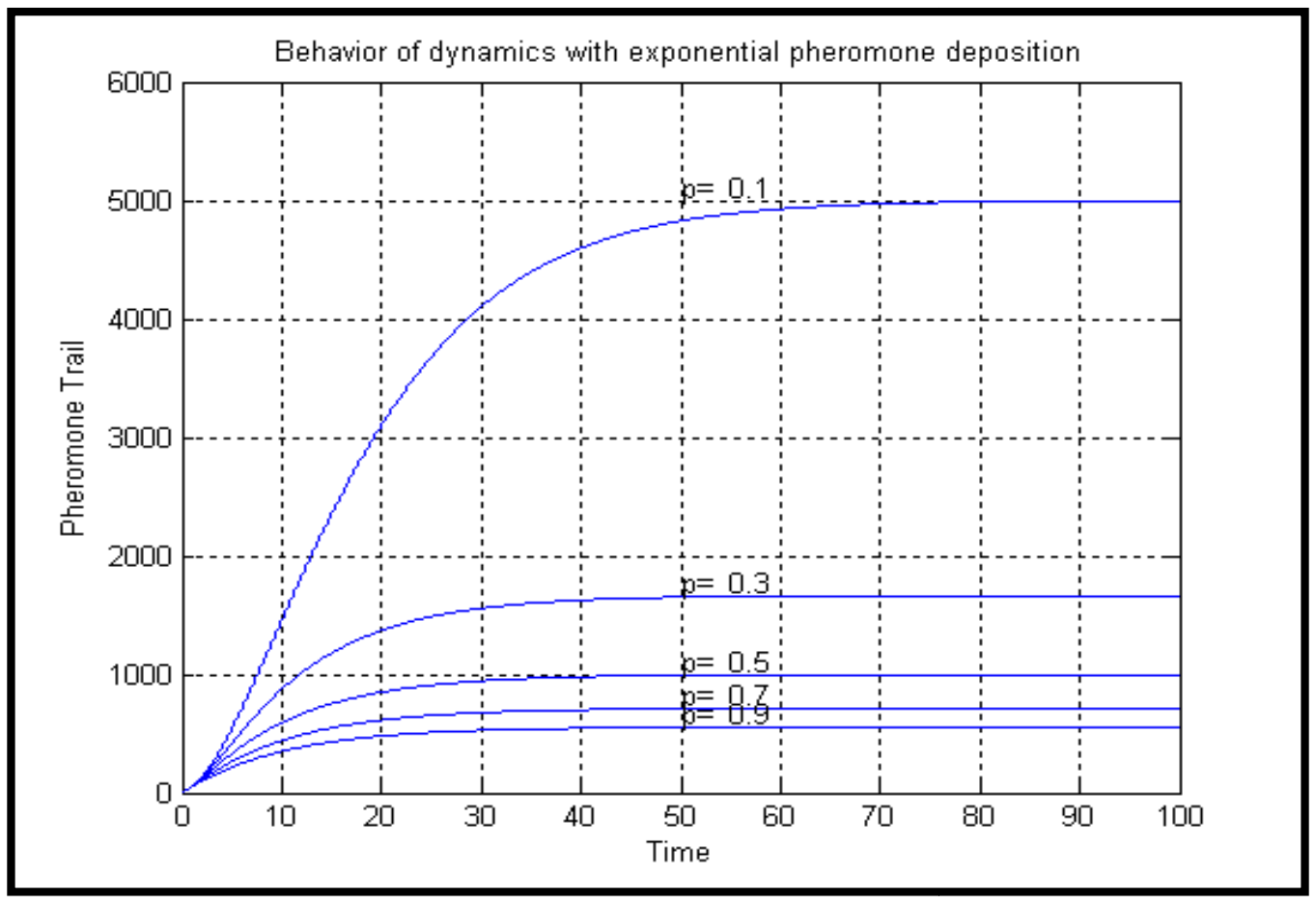

**Figure 3: $\tau_{ij}(t)$ versus t for exponential pheromone deposition with T=15**

## V. Parameter Values in MMAS

The main controlling parameters of MMAS algorithm are α, β, ρ (evaporation rate), $\tau_{min}$ (lower pheromone trail limit) and $\tau_{max}$ (upper pheromone trail limit). As suggested in [6], $\tau_{max}=1/\rho C^{bs}$ gives best result where $C^{bs}$ is the length of the best tour found in current iteration. Also if $p_{dec}$ be the probability of choosing a particular solution component at a choice point and an ant has to make $n$ successive right choices to construct the best solution, then the probability of selecting the best tour can be described as $p_{best}= p_{dec}^{\ n}$. In [6], it has been shown that $p_{best}$ relates $\tau_{min}$ and $\tau_{max}$ through the equation

$$\tau_{min} = \tau_{max}(1-\sqrt[n]{p_{best}})\,/\,[(avg-1)\sqrt[n]{p_{best}}\,] \quad (11)$$

where $avg$ is the average number of choices available to an ant at each step while constructing a solution. [6] shows that best performance in context of TSP with traditional MMAS algorithm is achieved with α=1.0, β=2.0, ρ=0.02, *m(*no of ants*)=N(*no of nodes*)* and $p_{best}$ =0.05. Our problem environment is different from TSP and therefore parameters, whose values depend on TSP instances, have to be modified. The number of successive right choices $n$ is assigned $N$ in case of TSP. But in our problem, optimum solution generated reveals that an ant has to make an average of 20 right decisions to reach its destination. The parameter $n$ is therefore set at 20. Consequently, value of $\tau_{min}$ gets altered. Value of $avg$ is set at 10 as it is found empirically that an ant has to choose on average of 10 paths at each step. Values of other parameters (α, β, ρ and $\tau_{max}$) are left unaltered. Also, in first 20 iterations, only iteration best solution is reinforced and afterwards reinforcement is given to best-so-far solution and iteration-best solution in alternate iteration. This strategy allows better exploration of search space as exemplified in [6].

## VI. Simulation Results

Max-Min Ant System (**MMAS**) model is considered here to study its performance with two kinds of deposition rules. As a problem environment, we take a network of connected cities where the shortest route between two given cities is to be determined. Now, suppose we have a starting city and a terminal city in a roadmap. Ants begin their tour at the starting city and terminate their journey at the destination city. Ant decides its next position at each intermediate step by a probability based selection approach. Suppose the neighborhood of an ant k currently residing on $i^{th}$ city be denoted by $\mathrm{N_i^k}$. Then the ant's choice of a city from its neighborhood $\mathrm{N_i^k}$ is governed by (1) with $P_i^k(j)$ as the probability of selecting city $j$ when the $k^{th}$ ant is in city $i$. A city $j \in \mathrm{N_i^k}$ if $j$ has not already been visited by $k^{th}$ ant. $\eta_{il}$ is the visibility information defined here as $\eta_{il} = 1/(|d_{il}|+|d_{lg}|)$ where $\boldsymbol{d_{il}}$ is defined as the distance between the cities $i$ and $l$ and $\boldsymbol{d_{lg}}$ specifies the distance between cities $l$ and $g$, $g$ being the destination city. **α, β** are the weights for pheromone concentration and visibility as usual. Ants stop moving if either they find a dead end or reach the destination city.

Constant deposition is the standard form of pheromone updating approach applied so far in all variants of ant system algorithms. In this approach, deposition of excess pheromone in all links of the tour is kept constant. But in our approach, we gradually increase the pheromone deposition on links as we move closer to the destination city. It implies that the links lying closer to the destination city receive more amount of pheromone as compared to those near the starting city.

We now present the results of experiments performed. We divide the simulation strategy in two levels. In the primary level, the two competitive algorithms are run on 20 different city distributions and the range of values of parameters of the proposed algorithm for which it performs best and outperforms its classical counterpart by largest extent is estimated. In section A, we tabulate results for only 3 out of those 20 different distributions.

**Section A:**
**Results for Roadmap I:**

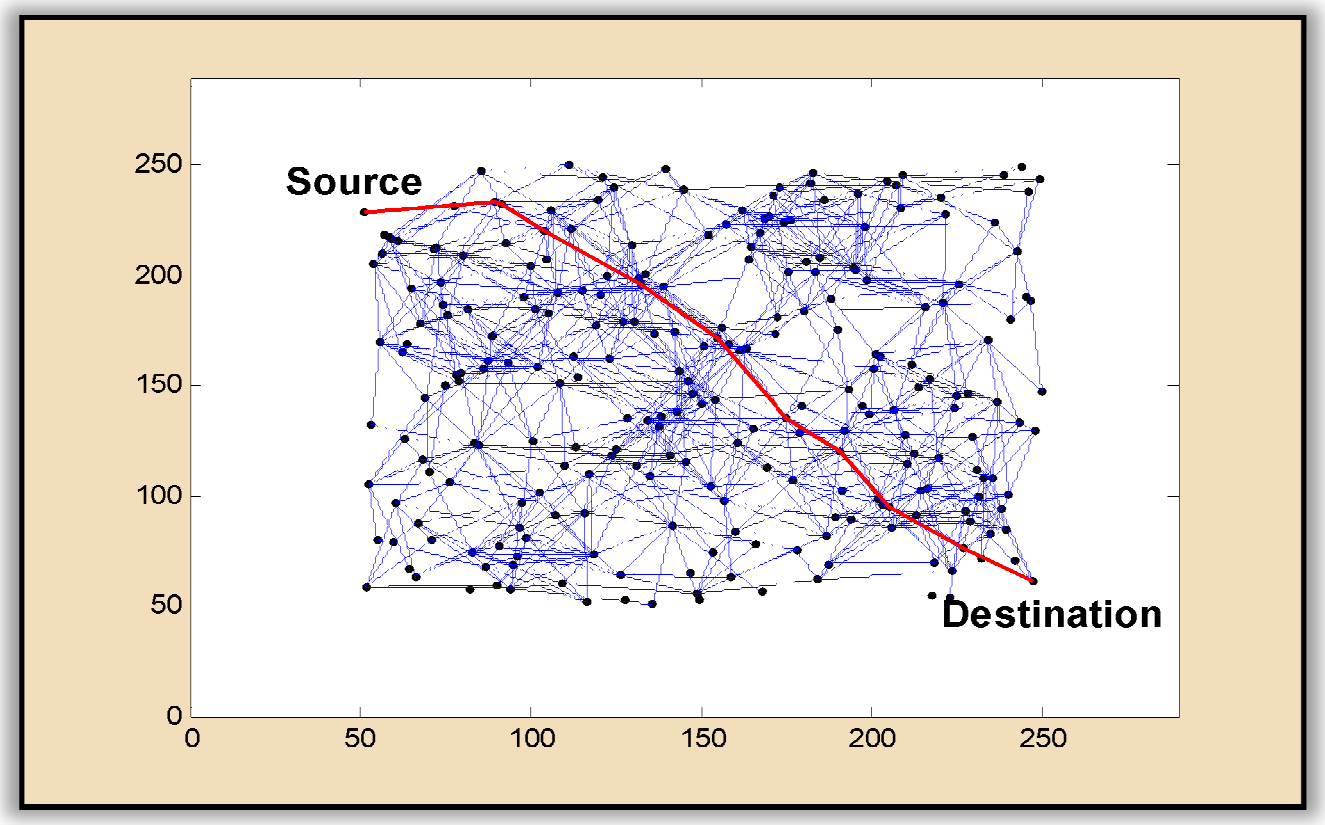

**Figure 4 : Roadmap for 250 City Distribution**

A sufficiently complex roadmap of 250 cities is taken as the first problem environment. Here, 20 ants are employed to move through the graph for 100 iterations to find out the optimal path length between the source and destination cities as highlighted in figure 4. Parameters **α** and **β** are both varied over the range 0.5 to 5.0 in steps of 0.5 to find out the optimum setting for which the algorithm gives best result. For roadmap I, optimum performance is achieved at α=0.5 and β=2.5. The best path found for this parameter setting closely matches with the theoretical minima as obtained from ***Dijkstra's algorithm*** ([14]). This theoretical minimum path is marked by red line in figure 4. Convergence time of the algorithm is defined as the minimum number of iterations required to converge to the optimum path. Evidently, even with respect to convergence time, the algorithm performs best for α=0.5, β=2.5. 3D plots of variations of optimum path length as well as convergence time for varying α, β are provided in figure 5 and 6. The location of minima at α=0.5, β=2.5 is obvious from the plot.

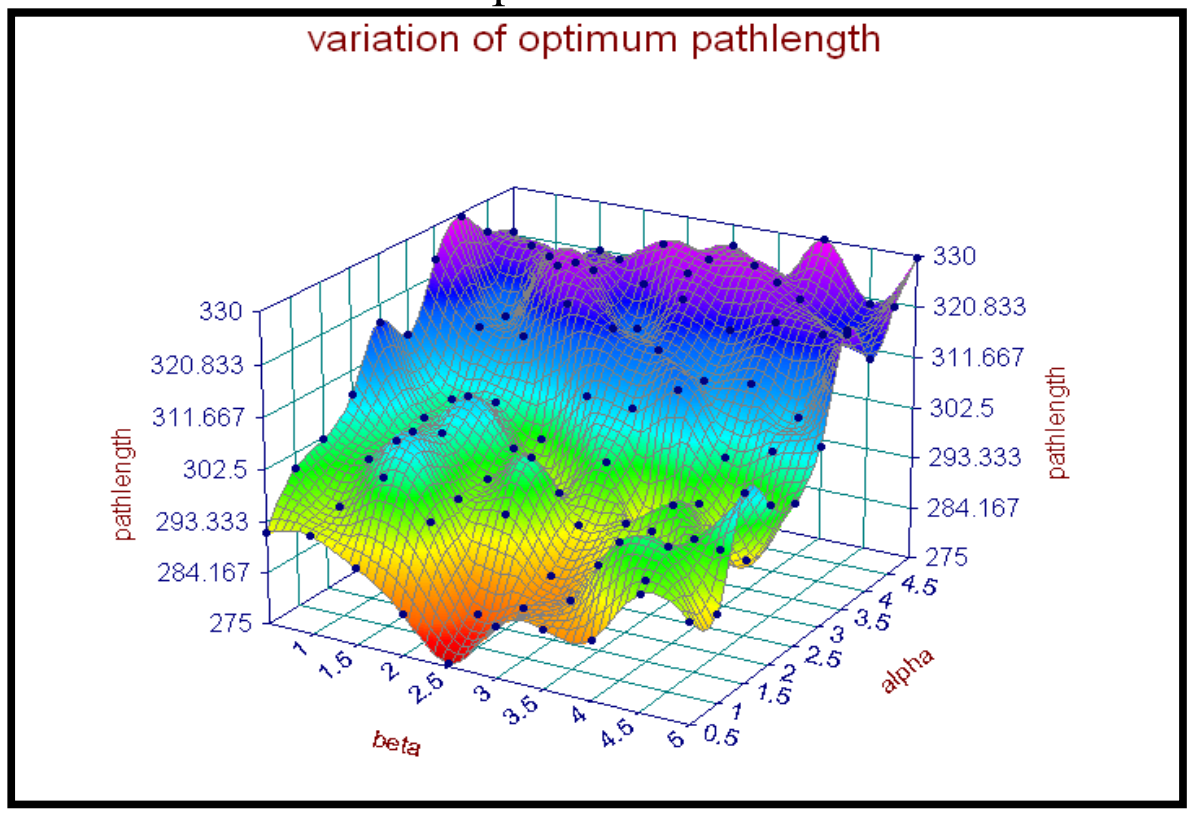

**Figure 5 : Variation of optimum path length with α and β for 250 city distribution**

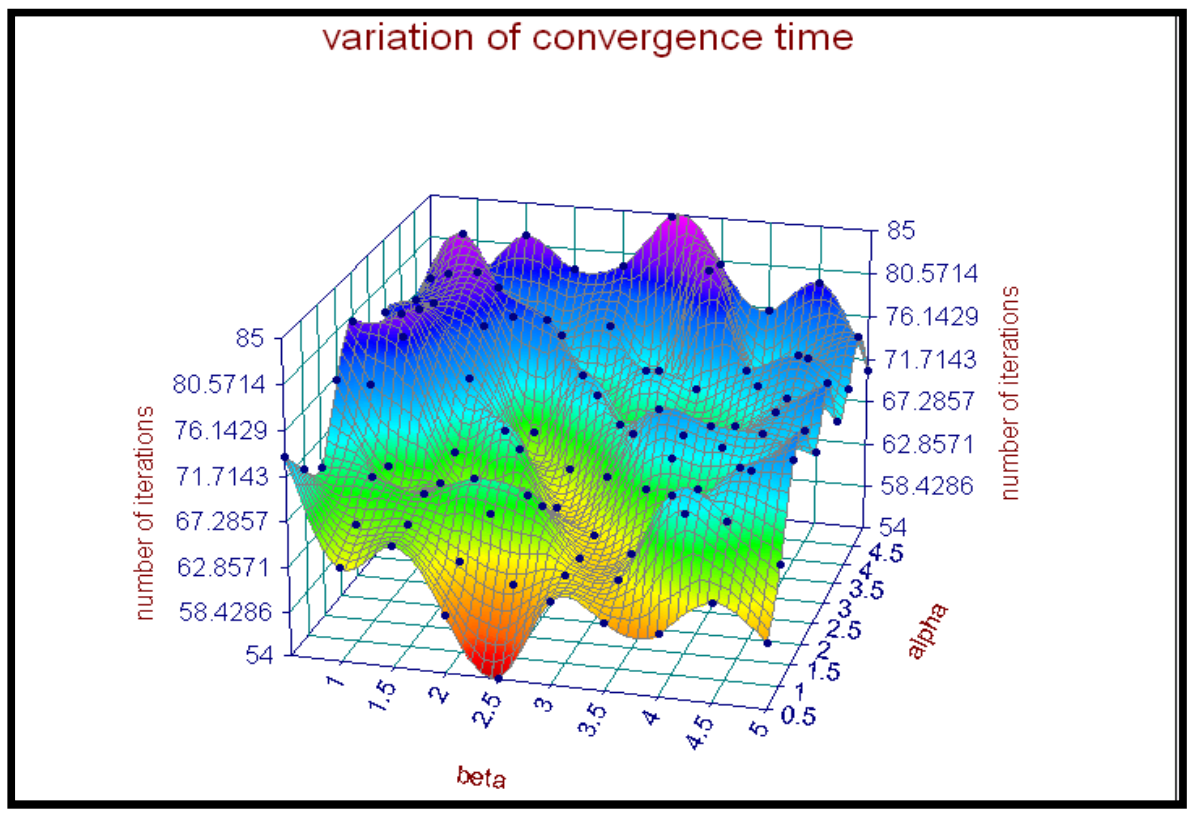

**Figure 6 : Variation of convergence time with α and β for 250 city distribution**

In all simulations above, parameter T is kept constant at a value 15. This value of T is guessed from the number of links required to move from source city to destination city. In most optimal solutions, number of links lies between 17 and 20. Hence, as far as the philosophy of exponential deposition rule is considered this value is quite convincing and that is also proved from the performance of the algorithm.

With optimum settings of controlling parameters, the newly proposed algorithm is next compared with the traditional MMAS algorithm. The betterment in both solution quality and convergence time is obvious from the plots of figure (7). The blue line marks the best-so-far path-length with uniform deposition approach and the red line marks the same with exponential deposition approach. The green line shows the theoretical minimum path between the source and destination cities.

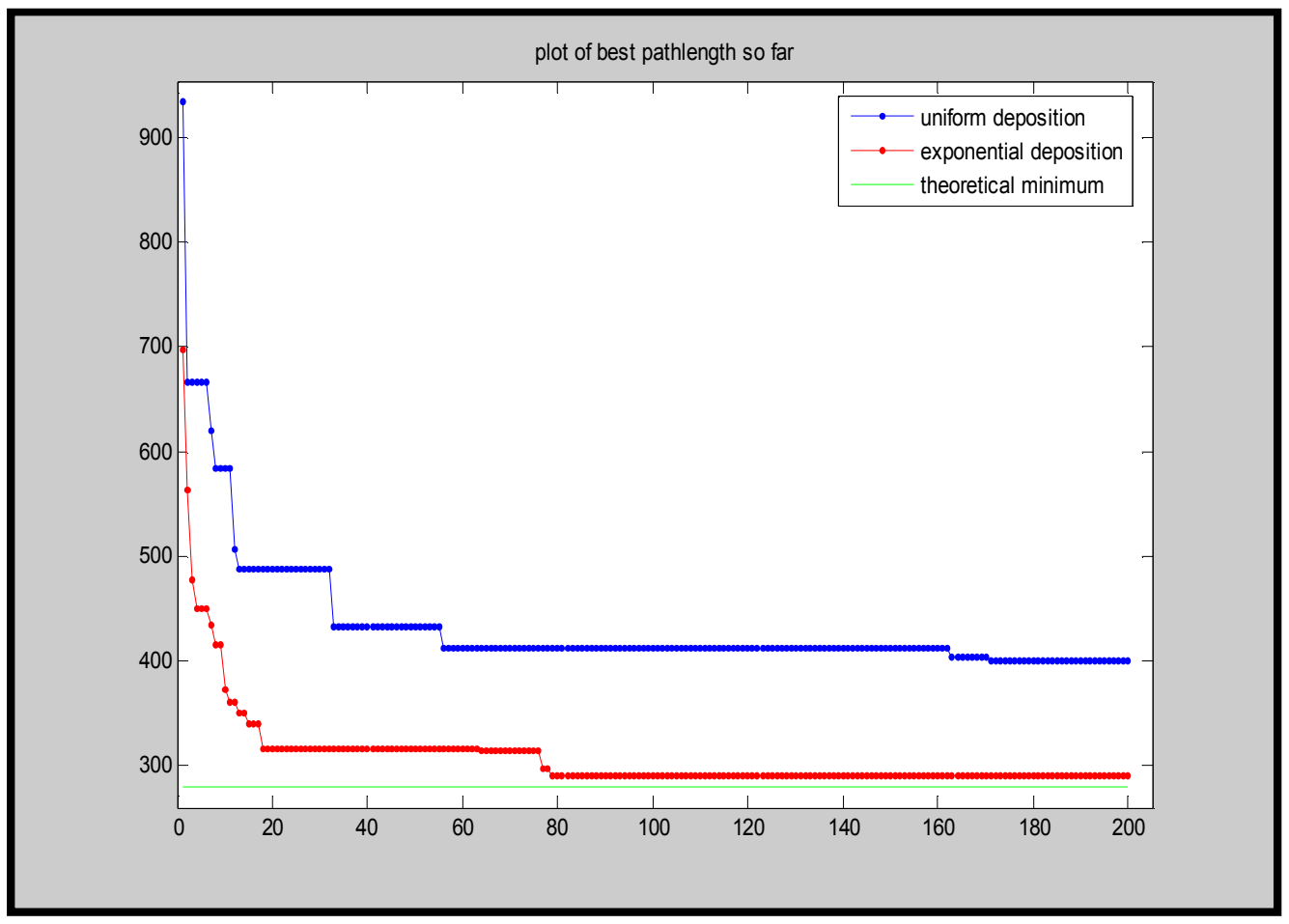

**Figure 7: Comparative study of algorithms**

## Results for Roadmap II:

Roadmap II is somewhat more complicated environment with 300 cities. Results reveal that for this environment best performance is achieved at **α=0.5** and **β=2.5**.

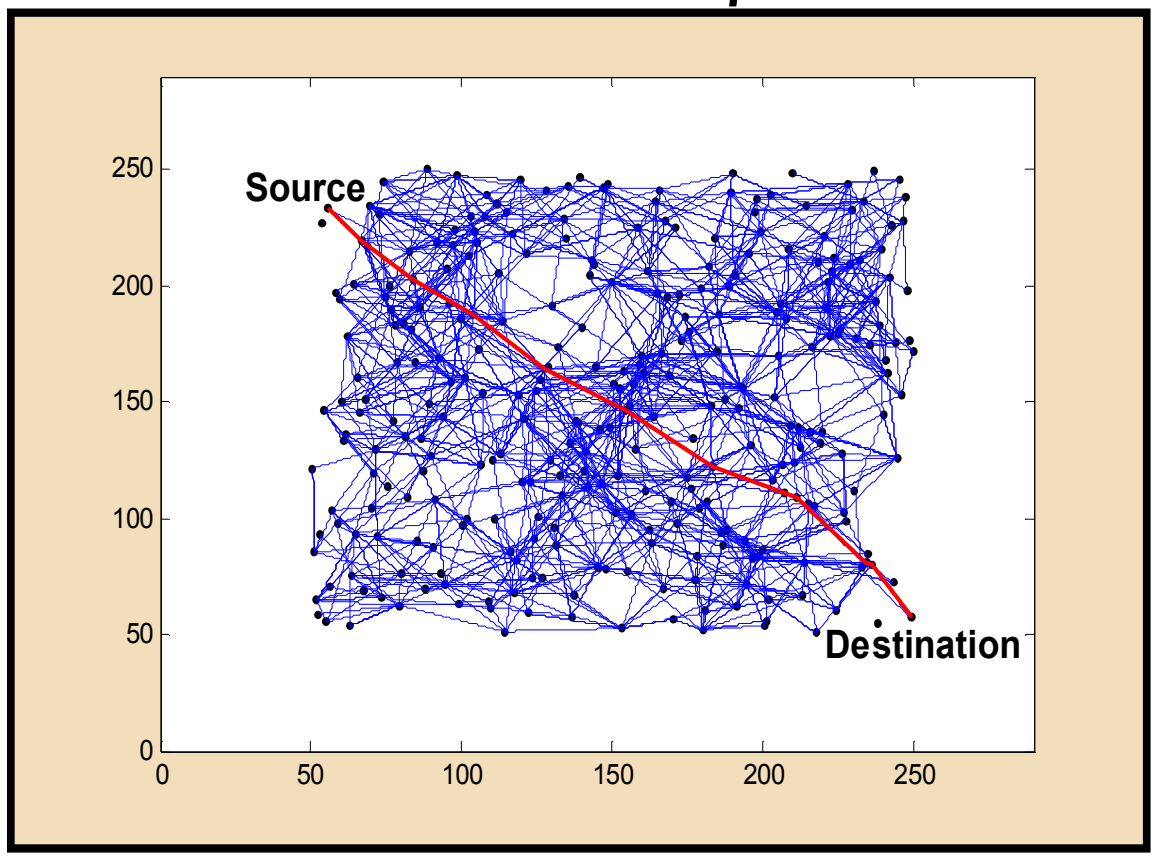

**Figure 8: Roadmap for 300 City Distribution**

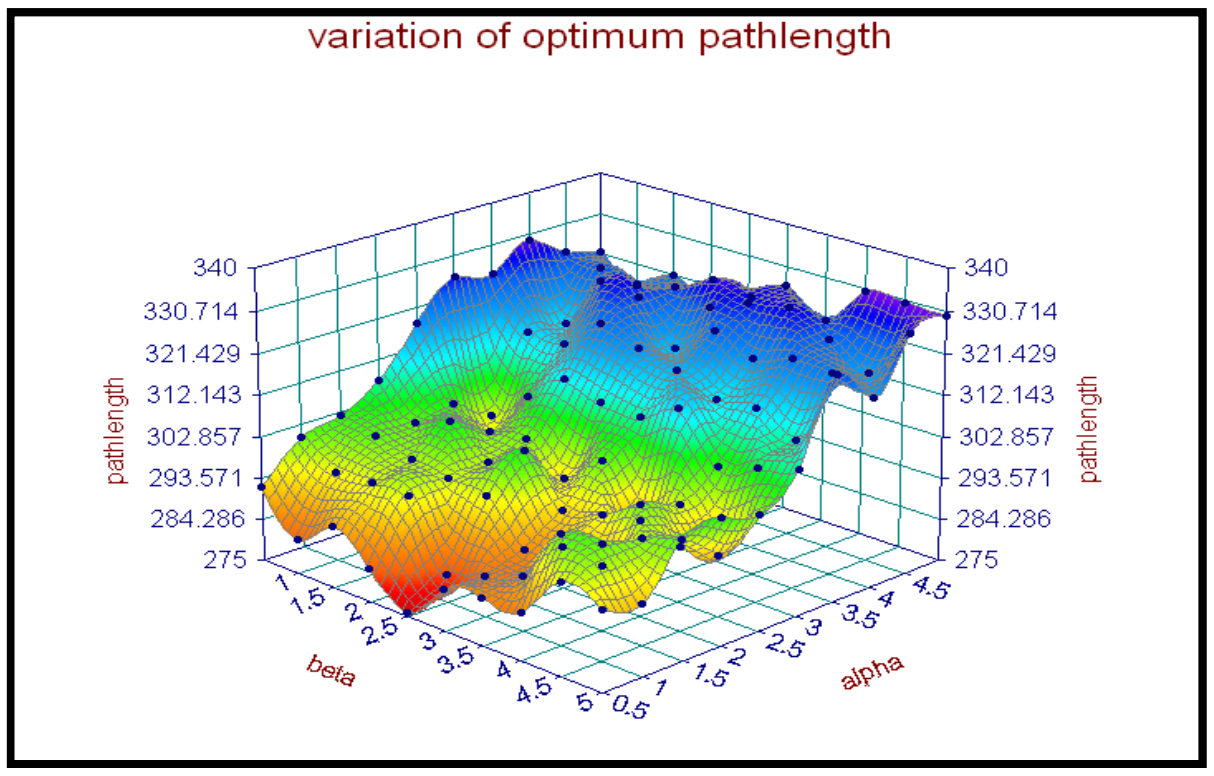

**Figure 9 : Variation of optimum path length with α and β for 300 city distribution**

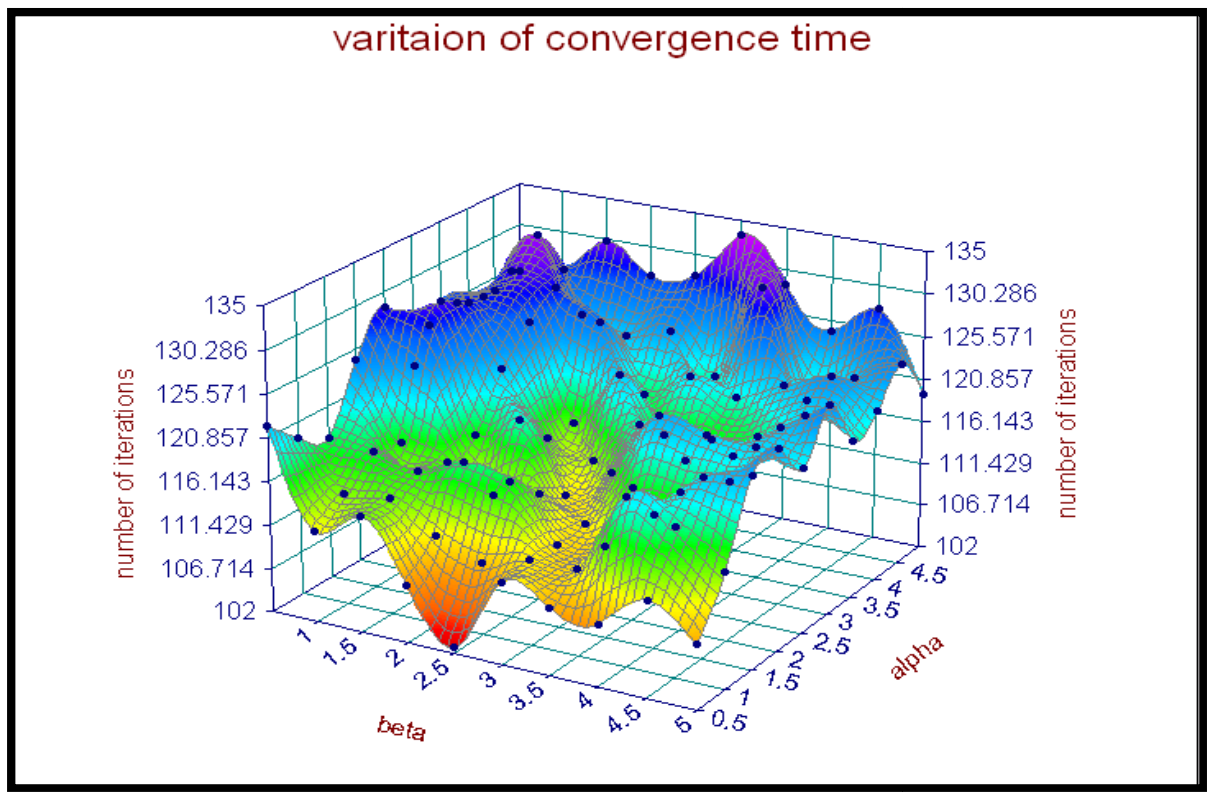

**Figure 10 : Variation of convergence time with α and β for 300 city distribution**

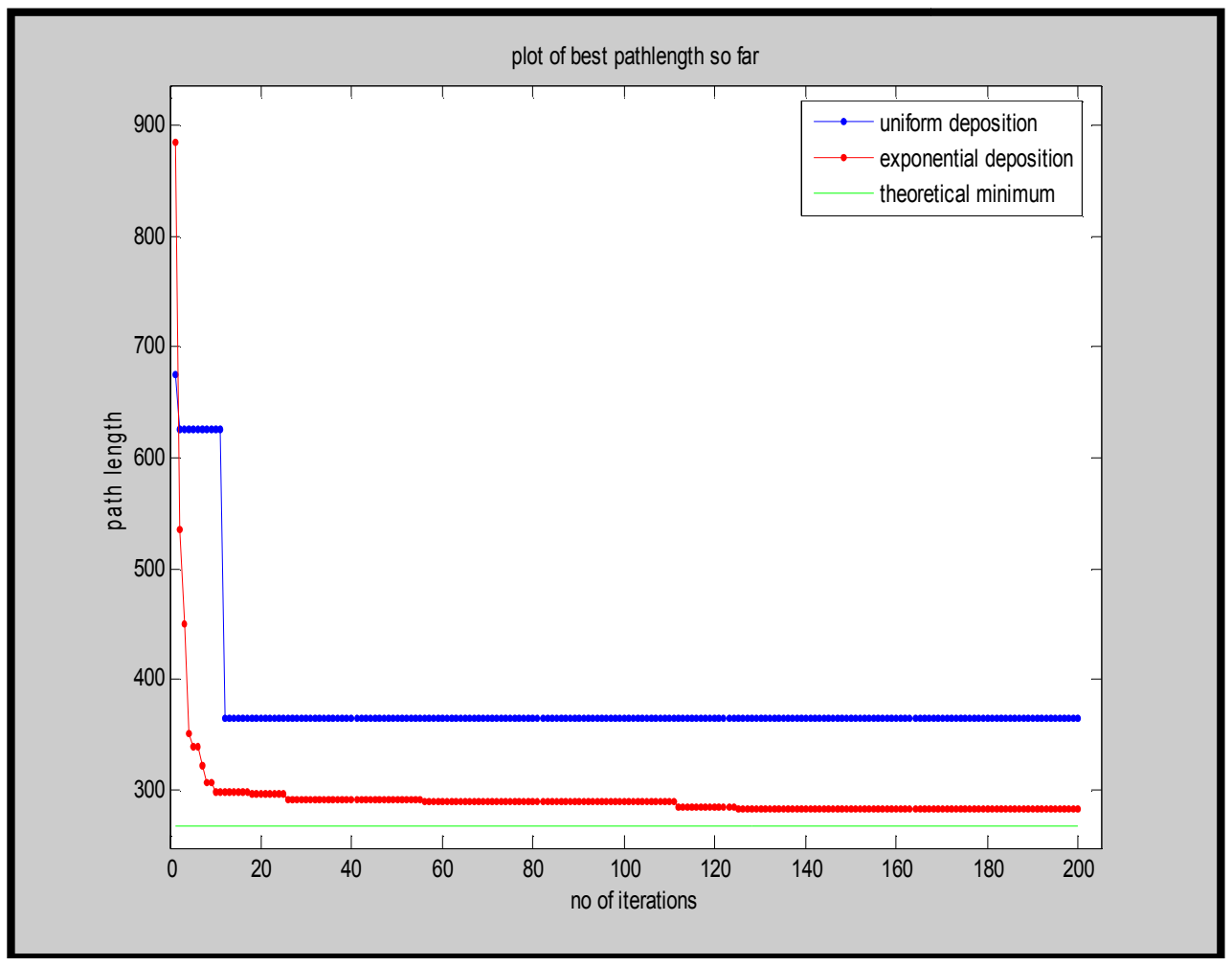

**Figure 11 : Comaparative Study of algorithms**

## Results for Roadmap III:

Optimum parameter setting : **α=0.5,β=3.0**.

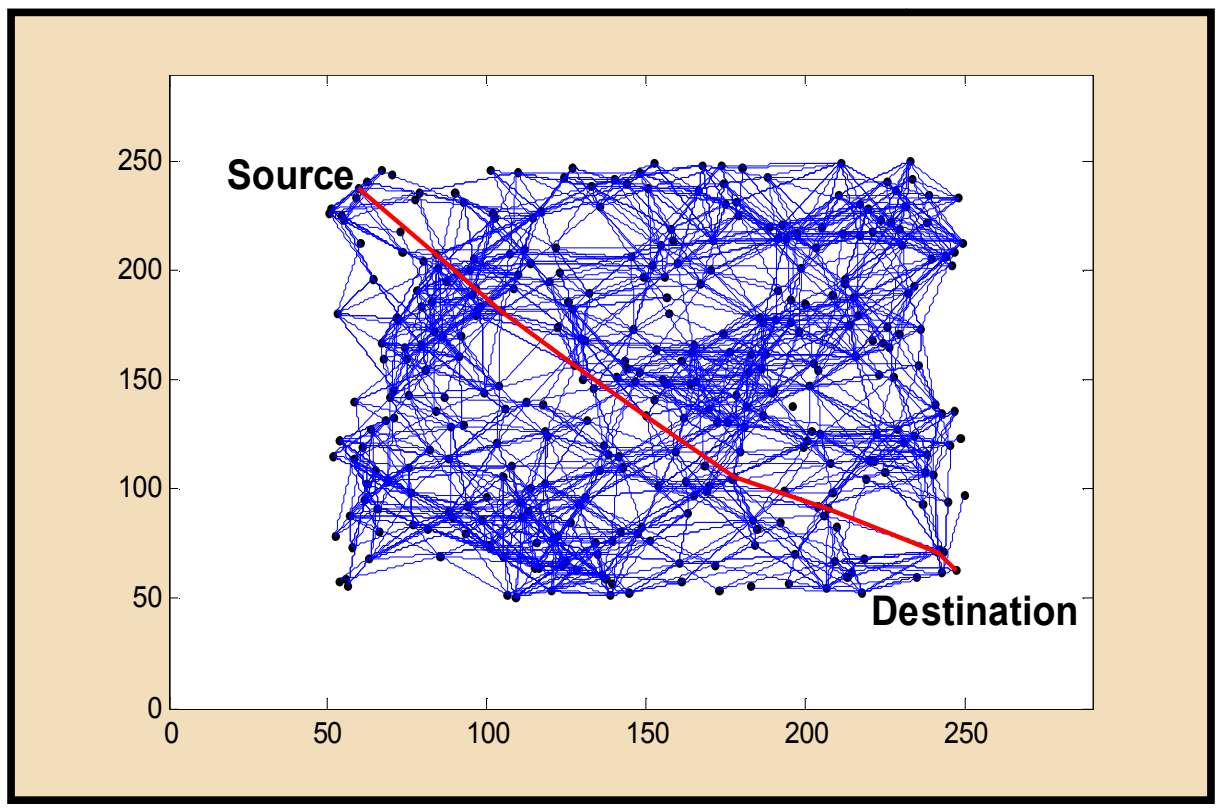

**Figure 12: Roadmap for 350 City Distribution**

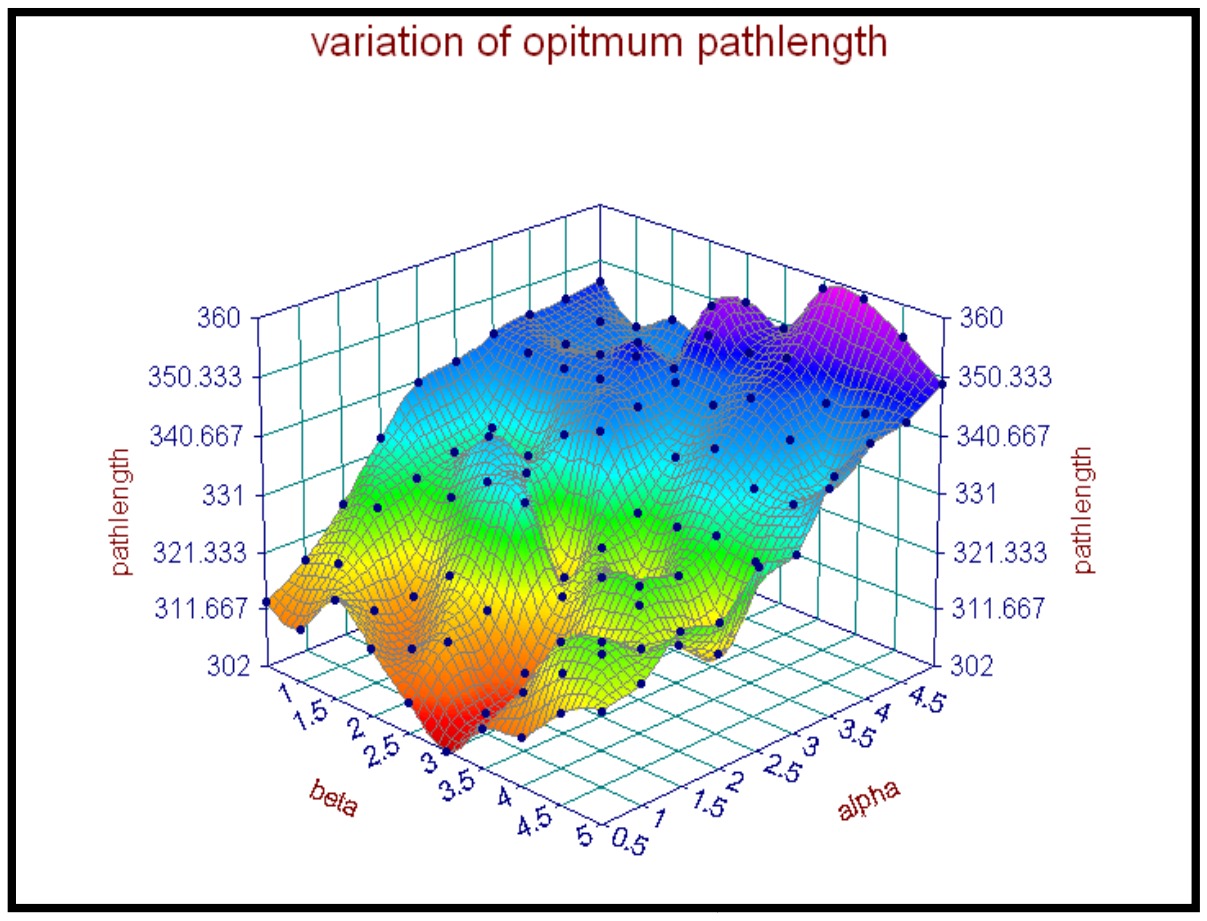

**Figure 13 : Variation of optimum path length with α and β for 300 city distribution**

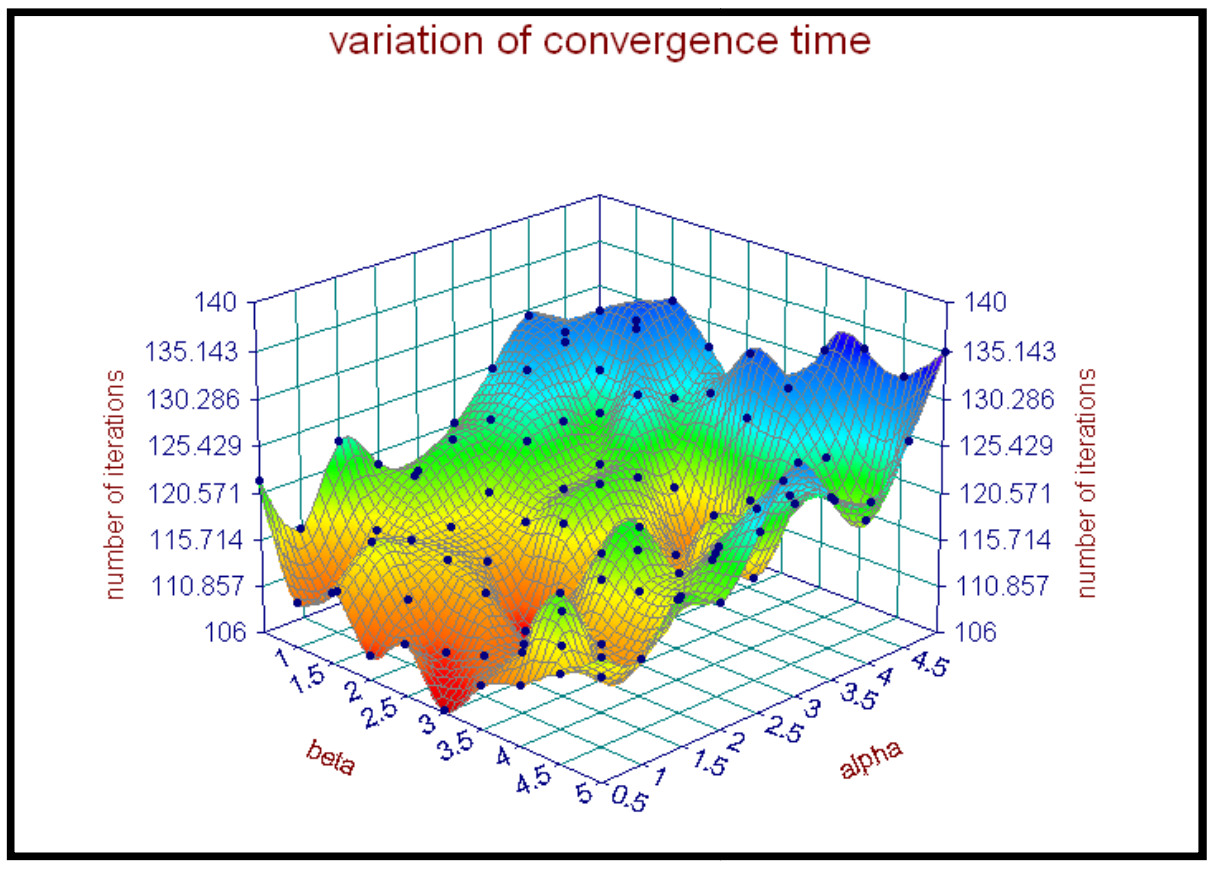

**Figure 14 : Variation of convergence time with α and β for 300 city distribution**

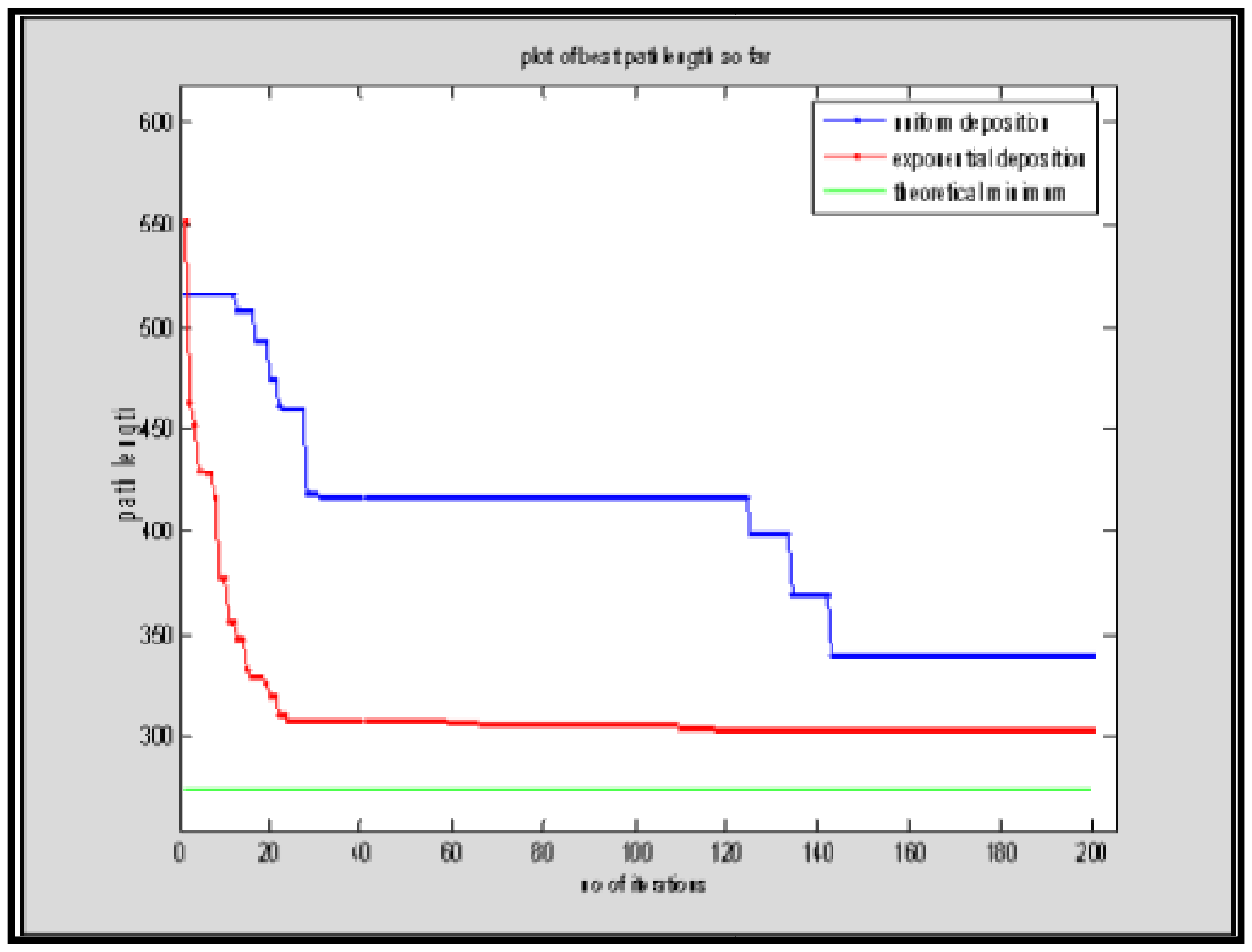

**Figure 15: Comaparative Study of algorithms**

## Section B:

Experimental results in section A reveal that the proposed algorithm performs best for α=0.5 and β lying between 2.5

and 3.0, no matter how complex the environment is. In secondary level of our simulation strategy, we vary α in the range of 0.25 to 0.75(i.e in the neighborhood of 0.5) and β over the range 2.5 and 3.0 in steps of 0.1 and try to estimate their relation with two features of problem environment: i) the ***node density*** and ii) ***standard deviation*** of lengths of smallest arc associated with each node. We performed experiments on roadmaps with 250, 265, 280, 295, 310, 325 and 350 number of cities. For each of above roadmaps, we choose seven different distributions and recorded the values of α and β for best performance of our algorithm. ***Table Curve 3D V4.0***, a curve fitting tool, is then employed to fit a curve through 49 data points for each of α and β and obtain an algebraic relation between α or β and the features of problem environment. The results are displayed in figures 16 and 17 along with the equations that relate the two sets of parameters. This exhaustive experimentation allows determination of optimum values of α and β when the features of problem environment are known in advance.

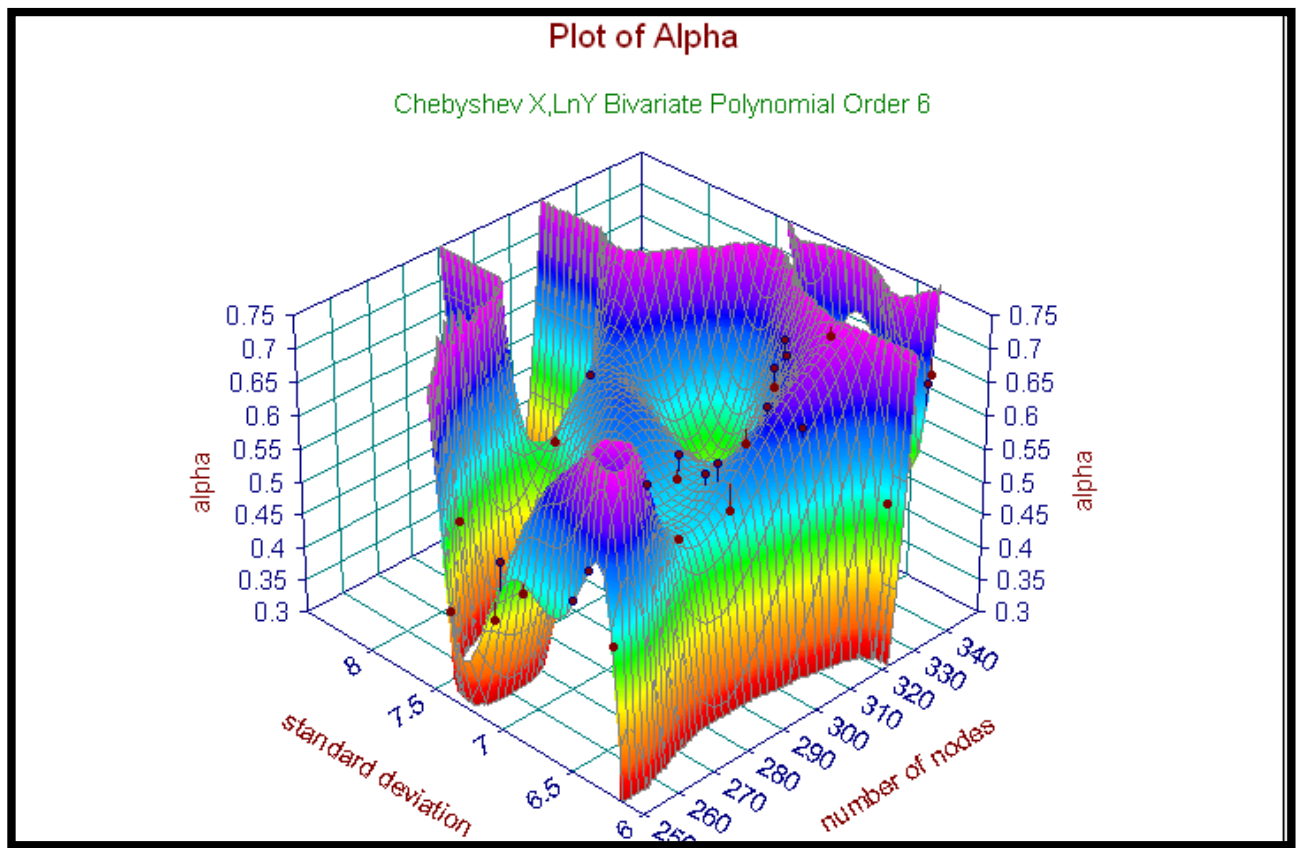

**Figure 16: Plot of α**

**Function: Chebyshev Series X,LnY Bivariate Polynomial Order 6**
**[x': x scaled -1 to +1; y': ln(y) scaled -1 to +1;**
**Tn(x') = cos(nacos(x'))**
**x ≡ number of nodes in 200 sq unit area; y ≡ standard deviation]**
**α=a+bT1(x')+cT1(y')+dT2(x')+eT1(x')T1(y')+fT2(y')+gT3(x')+hT2(x')T1(y')+iT1(x')T2(y')+jT3(y')+kT4(x')+lT3(x')T1(y')+mT2(x')T2(y')+nT1(x')T3(y')+oT4(y')+ pT5(x')+qT4(x')T1(y')+rT3(x')T2(y')+sT2(x')T3(y')+ tT1(x')T4(y')+uT5(y')+vT6(x')+aaT5(x')T1(y')+ abT4(x')T2(y')+acT3(x')T3(y')+adT2(x')T4(y')+ aeT1(x')+T5(y')+afT6(y')**
**Co-efficient values:**
**a=2.094, b=-5.892, c=-3.756,d=1.813, e= -8.864, f=1.257, g= 0.697, h= 2.269, i= 2.720, j=0.1556, k=0.9132, l=1.722, m= -1.423, n=0.232, o=0.743, p=1.270, q=1.345, r=0.412, s=0.738, t=1.575, u=0.774, v=0.604, aa=1.323, ab= -0.932, ac=-0.975, ad=-0.827,ae=-0.115, af=0.124**

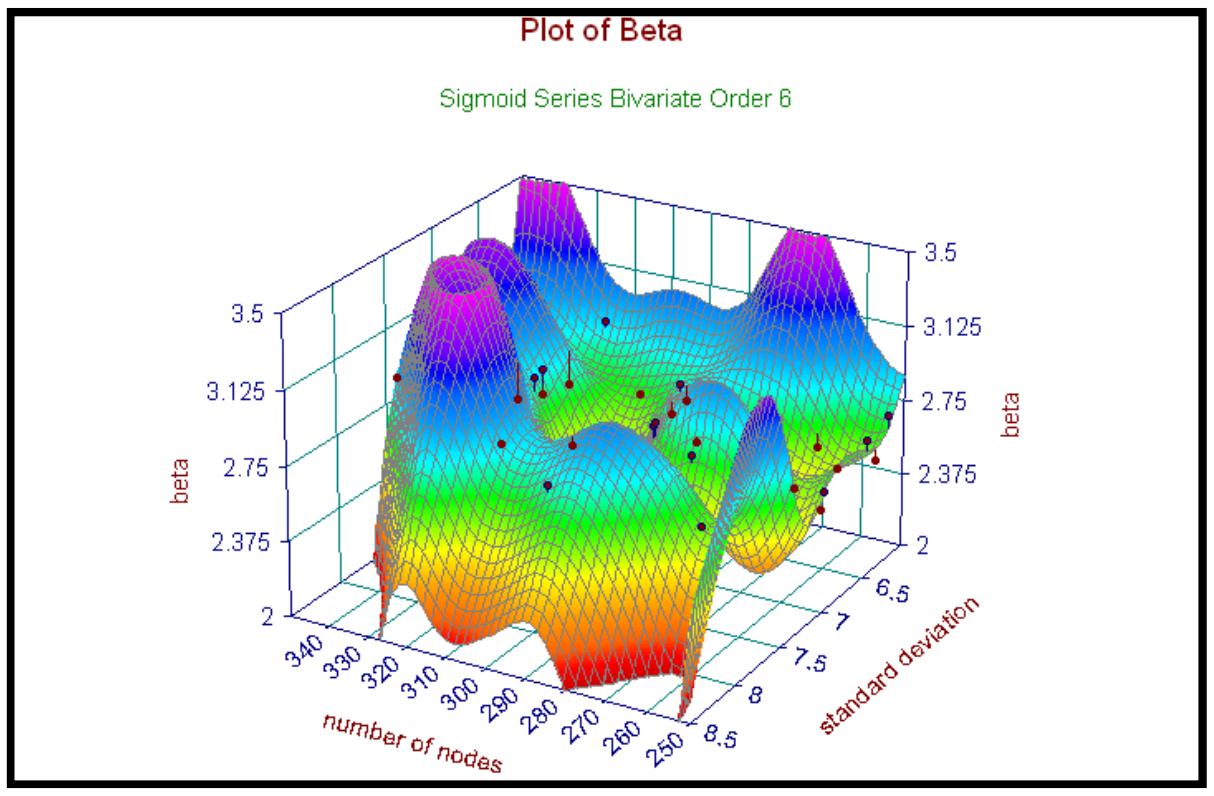

**Figure 17: Plot of β**

**Function: Sigmoid Series Bivariate Polynomial Order 6**
**[ x': x scaled -1 to +1; y': y scaled -1 to +1;**
**Si=2..n(x') = -1+2/(1+exp(-(x'+1-(i-1)*(2/n))/0.12)), S1(x') = x']**
**β=a+bS1(x')+cS1(y')+dS2(x')+eS1(x')S1(y')+fS2(y')+gS3(x')+hS2(x')S1(y')+iS1(x')S2(y')+jS3(y')+ kS4(x')+lS3(x')S1(y')+mS2(x')S2(y')+nS1(x')S3(y')+oS4(y') +pS5(x')+qS4(x')S1(y')+rS3(x')S2(y')+ sS2(x')S3(y')+tS1(x')S4(y')+uS5(y')+vS6(x')+ aaS5(x')S1(y')+abS4(x')S2(y')+acS3(x')S3(y')+ adS2(x')S4(y')+aeS1(x')S5(y')+afS6(y')**
**Co-efficient values:**
**a=1.396, b=-0.106, c=1.427, d=1.203, e=-1.107, f=1.115, g=-0.214, h=1.415, i=-0.413, j=0.753, k=0.116, l=-0.360, m=1.666 , n=-0.867, o=0.293 p=-0.066, q=0.274, r=-0.229, s=0.572, t=-0.191 u=0.081, v=-0.047, aa=-0.158, ab=0.152, ac=-0.140, ad=0.502 , ae=-0.272 , af=-0.036**

## VII. CONCLUSIONS AND FUTURE WORK

The stability analysis and pheromone deposition approach presented in this paper are both entirely novel. The exponential deposition approach outperformed the classical one by a large margin and has lead to better solution quality and algorithm convergence. Our next venture includes studying the comparative behavior of the two kinds of deposition approach in other models of extended Ant System algorithm like the Rank-based Ant System, Ant Colony System and Elitist Ant System.